\documentclass[a4paper,fleqn]{cas-dc}
\usepackage{amssymb}
\usepackage{hyperref}
\usepackage{graphicx}
\usepackage{lipsum}
\usepackage{lscape}
\usepackage{amsmath}
\usepackage{relsize}%
\usepackage{multirow}
\usepackage[figuresright]{rotating}
\usepackage{lineno}
\usepackage{algorithm} 
\usepackage{algpseudocode} 
\usepackage{comment}
\usepackage{lineno}
\usepackage{nomencl}
\makenomenclature
\usepackage{calc}
\usepackage[T1]{fontenc}
\usepackage{relsize}
\usepackage{subcaption}

\usepackage{natbib}
\setcitestyle{authoryear,open={(},close={)},citesep={;}} 

\hypersetup{
  colorlinks,
  citecolor=Violet,
  linkcolor=Red,
  urlcolor=Blue}

\newcolumntype{P}[1]{>{\centering\arraybackslash}p{#1}}

\def\tsc#1{\csdef{#1}{\textsc{\lowercase{#1}}\xspace}}
\tsc{WGM}
\tsc{QE}
\tsc{EP}
\tsc{PMS}
\tsc{BEC}
\tsc{DE}

\begin{document}
\let\WriteBookmarks\relax
\def\floatpagepagefraction{1}
\def\textpagefraction{.001}
\shorttitle{Semi-Supervised Weed Detection}
\shortauthors{Li et~al.}
 
\title [mode = title]{Performance Evaluation of Semi-supervised Learning Frameworks for Multi-Class Weed Detection}

\author[1]{Jiajia Li}\ead{lijiajia@msu.edu}
\author[2]{Dong Chen}\ead{chendon9@msu.edu}
\author[3]{Xunyuan Yin}\ead{xunyuan.yin@ntu.edu.sg}
\author[1]{Zhaojian Li*}\ead{lizhaoj1@egr.msu.edu}

\address[label1]{Michigan State University, East Lansing, MI, USA}
\address[label2]{University of Virginia, Charlottesville, VA, USA}
\address[label3]{Nanyang Technological University, Singapore}
\address{* Zhaojian Li is the corresponding author}

\begin{abstract}
Effective weed control plays a crucial role in optimizing crop yield and enhancing agricultural product quality. However, the predominant reliance on herbicide application not only poses a critical threat to the environment but also promotes the emergence of resistant weeds. Fortunately, recent advances in precision weed management enabled by machine vision and deep learning provides a sustainable alternative. Despite great progresses, existing perception algorithms are mainly developed based on supervised learning approaches, which typically demands large-scale datasets with manual-labeled annotations, which is time-consuming and labor-intensive. 
As such, label-efficient learning methods, especially semi-supervised learning, have gained increased attention in the broader domain of computer vision and have demonstrated promising performance. These methods aim to utilize a small number of labeled data samples along with a great number of unlabeled samples to develop high-performing models comparable to the supervised learning counterpart trained on a large amount of labeled data samples. In this study, we assess the effectiveness of a semi-supervised learning framework for multi-class weed detection, employing two well-known object detection frameworks, namely FCOS and Faster-RCNN. Specifically, we evaluate a generalized student-teacher framework with an improved pseudo-label generation module to produce reliable pseudo-labels for the unlabeled data. To enhance generalization, an ensemble student network is employed to facilitate the training process.
Experimental results show that the proposed approach is able to achieve approximately 76\% and 96\% detection accuracy as the supervised methods with only 10\% of labeled data in CottenWeedDet3 and CottonWeedDet12, respectively. We offer access to the source code (\url{https://github.com/JiajiaLi04/SemiWeeds}), contributing a valuable resource for ongoing semi-supervised learning research in weed detection and beyond.
\end{abstract}

\maketitle
\section{Introduction}
\label{sec:intro}
Weeds pose a significant risk to global crop production, with potential losses attributed to these unwelcome plants estimated at 43\% worldwide \citep{oerke2006crop}. Specifically, in the context of cotton farming, inefficient management of weeds can result in a staggering 90\% reduction in yield \citep{manalil2017weed}. Traditional weed control methods typically involve the use of machinery, manual weeding, or application of herbicides. These weed management approaches, while commonly utilized, require significant labor and cost considerations. Manual and mechanical weeding methods are especially labor-intensive, a predicament that has been intensified by recent global labor shortages triggered by public health crises (e.g., the COVID-19 pandemic) and geopolitical conflicts (e.g., the Russia-Ukraine War) \citep{laborde2020covid, ben2022impacts}. Furthermore, the use of herbicides brings about significant environmental harm and potential risks to human health, and contributes to the emergence of herbicide-resistant weed species \citep{norsworthy2012reducing, chen2022performance}.

Precision weed management (PWM), which incorporates sensors, computer systems, and robotics into farming systems, has risen as a viable and sustainable strategy for effective weed control \citep{young2013future}. It allows for targeted treatment based on specific site conditions and weed species, thereby significantly minimizing the use of herbicides and other resources \citep{gerhards2003real}. To achieve successful implementation of PWM, it is essential to accurately identify, localize, and monitor weeds, which requires robust machine vision algorithms for weed recognition \citep{chen2022performance}. 
Traditional image processing techniques, often encompassing edge detection, color analysis, and texture feature extraction, along with subsequent steps such as thresholding or supervised modeling, are widely utilized in the field of weed classification and detection \citep{meyer2008verification, wang2019review}. For instance, a weed classification algorithm that relies on extracted texture features was developed in \citep{bawden2017robot}.  \cite{ahmad2018visual} used local shape and edge orientation features to differentiate between monocot and dicot weeds. However, despite promising results, these conventional machine vision techniques often necessitate manual feature engineering for specific weed detection or classification tasks, which requires extensive domain knowledge and can be error-prone and time-consuming. Moreover, these methods may struggle with complex visual tasks and be sensitive to variations in lighting conditions and occlusion \citep{o2020deep}. 

Recently, deep learning (DL) based advanced computer vision has been recognized as a promising approach for sustainable weed management \citep{chen2022performance, rai2023applications, rahman2023performance, dang2023yoloweeds, coleman2023image}. For example, in \cite{chen2022performance}, 35 state-of-the-art deep neural networks (DNNs) were examined and benchmarked for multi-class weed classification within cotton production systems, with nearly all models attaining high classification accuracy, reflected by F1 scores exceeding 95\%. Furthermore, in \cite{rahman2023performance}, 13 deep learning-based object detectors were assessed for weed detection. RetinaNet (R101-FPN) \citep{lin2017focal} emerged as the top performer, boasting a mean average precision (mAP@0.50) of 79.98\%. Despite their proven effectiveness, these DL-based approaches are notoriously data-hungry, and their performance is heavily dependent on large-scale and accurately labeled image datasets \citep{lu2020survey, rai2023applications}, whereas manually labeling such large-scale image datasets is often error-prone, tedious, expensive, and time-consuming \citep{li2023label}.

To address these challenges, label-efficient learning algorithms \citep{li2023label} have emerged as promising solutions to reduce the high labeling costs by harnessing the potential of unlabeled samples. Specifically, in \cite{dos2019unsupervised}, the efficacy of two popular unsupervised learning algorithms, namely Joint Unsupervised Learning of Deep Representations and Image Clusters (JULE, \cite{yang2016joint}) and Deep Clustering for Unsupervised Learning of Visual Features (DeepCluster, \cite{caron2018deep}), were evaluated in the context of weed recognition utilizing two publicly available weed datasets.
In addition, the semi-supervised learning for weed classification was studied in \citep{liu2023semi, liu2024semi, benchallal2024convnext}. 
Furthermore, a semi-supervised learning strategy called SemiWeedNet was introduced in \cite{nong2022semi}; this method was designed for the segmentation of weeds and crops in challenging environments characterized by complex backgrounds. Moreover, the study presented in \cite{hu2021powerful} employed the cut-and-paste image synthesis approach and semi-supervised learning to address the issue of insufficient training data for weed detection. This approach was evaluated on an image dataset consisting of 500 images across four categories: ``cotton'', ``morningglory'', ``grass'', and ``other'', which culminated in an mAP of 46.0. Although the results were intriguing, their methodology was tested only on a two-stage object detector (i.e., Faster-RCNN \citep{ren2015faster}) and a four-category image dataset, which does not sufficiently substantiate the efficacy of semi-supervised learning for weed detection. Therefore, our research aims to further probe the potential of semi-supervised learning in weed detection, and comparatively assess a variety of object detectors and multi-class weed species. The key contributions of this study are as follows:
\begin{itemize}
    \item We rigorously evaluate the semi-supervised learning framework utilizing two open-source cotton weed datasets. These datasets include 3 and 12 weed classes commonly found in U.S. cotton production systems. 
    \item We further analyze and compare the performance of one-stage and two-stage object detectors within the semi-supervised learning framework.
    \item In the spirit of reproducibility, we make all our training and evaluation codes\footnote{\url{https://github.com/JiajiaLi04/SemiWeeds}} freely accessible. 
\end{itemize}

The remainder of this paper is organized as follows: Section \ref{sec: methods} details the dataset and technical aspects pertinent to this study. Section \ref{sec: results} presents experimental results and provides a comprehensive analysis, followed by further discussions and limitations in Section \ref{sec: disscussion}. Lastly, Section \ref{sec: conclu} offers concluding remarks and outlines potential future research directions.

\section{Materials and Methods} 
\label{sec: methods}
In this section, we begin by introducing the two datasets employed in our study. Then, we provide an overview of two representative object detectors: the two-stage Faster R-CNN and the one-stage FCOS detector, along with the details of our semi-supervised framework. Lastly, we present the evaluation metrics and describe the experimental setups. 

\subsection{Weed Datasets}
To assess the performance and efficacy of our semi-supervised framework, we conduct evaluations on two publicly available weed datasets tailored specifically to the U.S. cotton production systems: CottonWeedDet3 \citep{rahman2023performance} and CottonWeedDet12 \citep{dang2023yoloweeds}.

\begin{figure}[!ht]
  \centering
\includegraphics[width=0.45\textwidth]{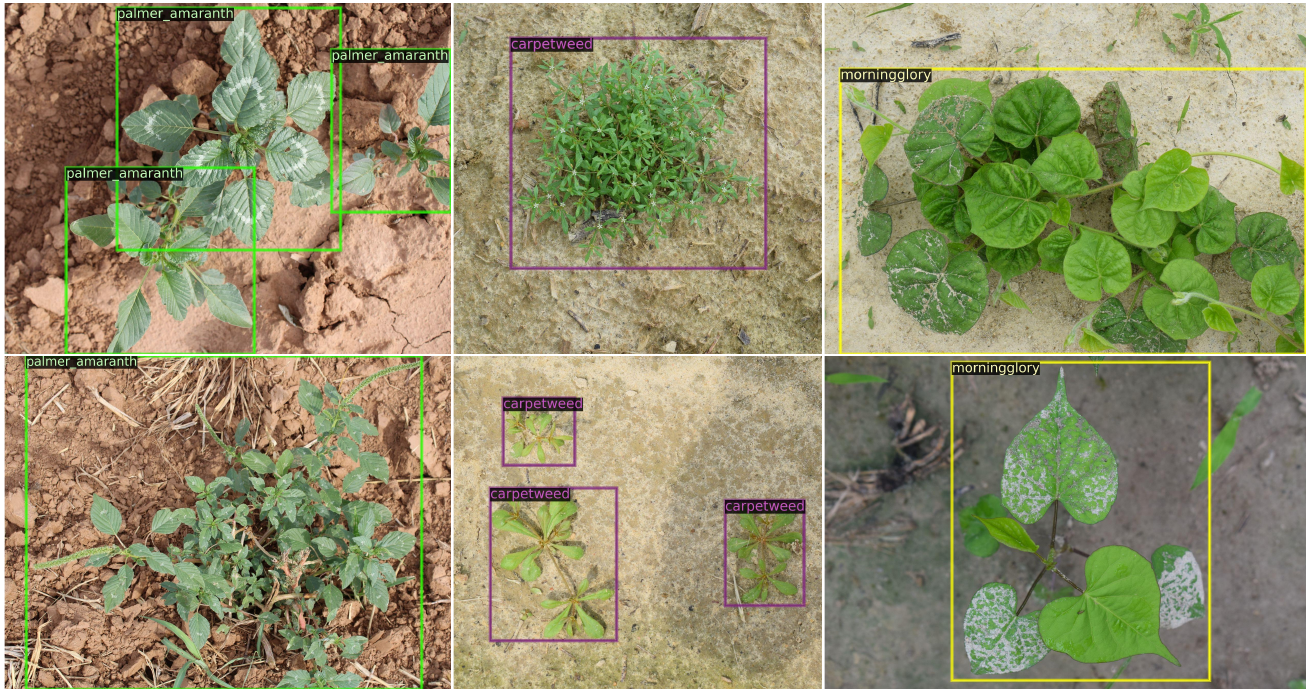}
  \caption{Weed samples in the CottonWeedDet3 dataset \citep{rahman2023performance}. Each column represents the image samples for one weed class.}
  \label{fig:CottonWeedDet3}
  \vspace{-10pt}
\end{figure}

CottonWeedDet3\footnote{CottonWeedDet3 dataset: \url{https://www.kaggle.com/datasets/yuzhenlu/cottonweeddet3}} \citep{rahman2023performance} comprises 848 high-resolution images ($4442 \times 4335$ pixels) annotated with 1532 bounding boxes. It contains three distinct classes of weeds commonly found in southern U.S. cotton fields, primarily in North Carolina and Mississippi. These images include three types of weeds: Carpetweed (Mollugo verticillata), Morningglory (Ipomoea genus), and Palmer Amaranth (Amaranthus palmeri). For adaptability, the annotations for each image are saved in both YOLO and COCO formats. Notably, around 99\% of the images contain less than 10 bounding boxes, with only a small portion (9 out of the 848 images) containing a more substantial quantity of bounding boxes, even up to 93 in some cases. Additionally, Carpetweed is the most frequently annotated, while Palmer Amaranth is the least. Visual examples of the three-class weed images can be found in Figure \ref{fig:CottonWeedDet3}.

\begin{figure*}[!ht]
  \centering
\includegraphics[width=0.95\textwidth]{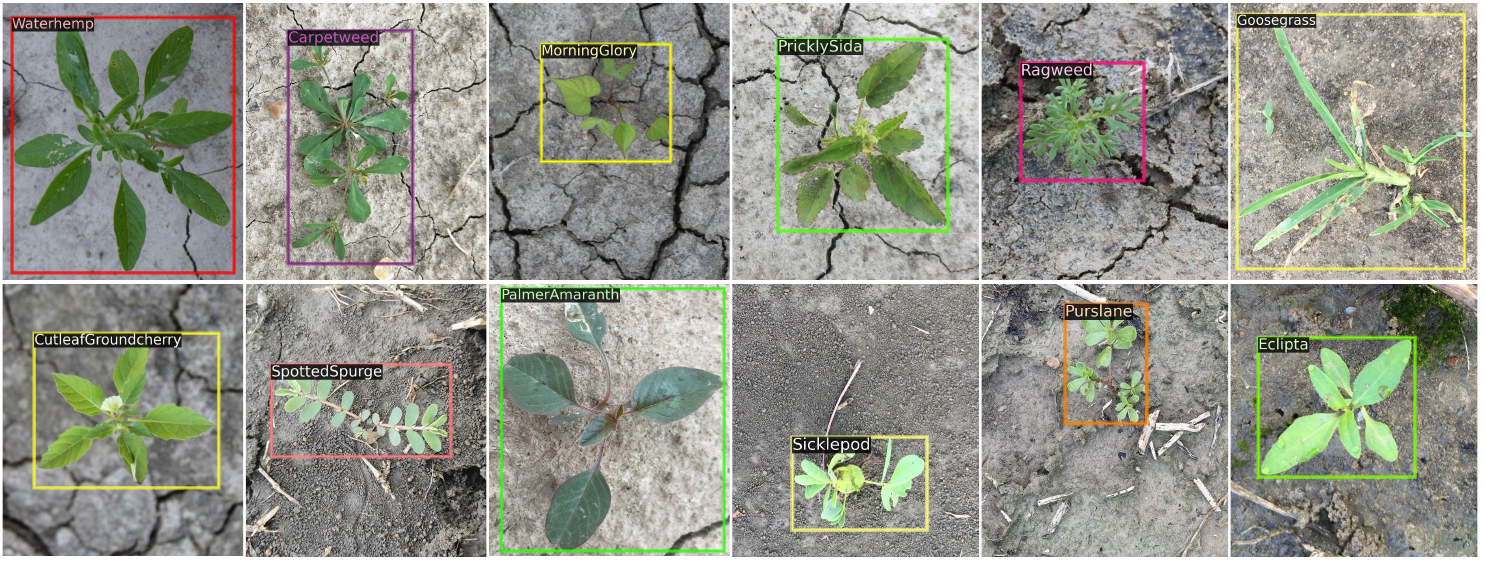}
  \caption{Weed samples in the CottonWeedDet12 dataset \citep{dang2023yoloweeds}.}
  \label{fig: CottonWeedID12 sample}
\end{figure*}

CottonWeedDet12 dataset\footnote{CottonWeedDet12 dataset: \url{https://zenodo.org/record/7535814}} \citep{dang2023yoloweeds} contains 5648 images of 12 weed classes, annotated with a total of 9370 bounding boxes (saved in both YOLO and COCO formats). These images, with a resolution exceeding 10 megapixels, were captured under natural lighting conditions and across various weed growth stages in cotton fields. Each weed class is represented by more than 140 bounding boxes.  Moreover, Waterhemp and Morning Glory have the highest number of bounding boxes while Googe Grase and Cutleaf Ground Cherry have the least.  In terms of image volume, the CottonWeedDet12 dataset surpasses the CottonWeedDet3 dataset \citep{rahman2023performance} by more than tenfold. Moreover, it represents the most extensive public dataset currently available for weed detection in cotton production systems. Figure \ref{fig: CottonWeedID12 sample} shows sample annotated images where a single weed class in each image is present, despite that each image may include multiple weed classes in the dataset.

\subsection{DL-based Object Detectors}
Deep Learning (DL)-based object detectors are typically structured around two primary components: a backbone and a detection head \citep{bochkovskiy2020yolov4}. The backbone is responsible for extracting features from high-dimensional inputs and is commonly pre-trained on ImageNet data \citep{deng2009imagenet}. Conversely, the head is leveraged to predict the classes and bounding boxes of objects. Existing detectors consist of anchor-based detectors \citep{ren2015faster, cai2016unified, lin2017focal} and anchor-free detectors \citep{law2018cornernet, tian2019fcos, zhou2019bottom}. Anchor-based detectors utilize pre-defined anchor boxes, adjusting them for position shifts and scaling to align with the ground-truth boxes, primarily based on their intersection-over-union (IoU) scores. Conversely, the pre-defined anchor boxes are discarded in the detection head for the anchor-free object detection models.

\subsubsection{Anchor-based detectors}
Anchor-based object detectors utilize pre-defined anchor boxes to efficiently localize and classify objects in images, being a representative approach in object detection methodologies. These methods have led to significant advancements and impressive outcomes in object detection \citep{ren2015faster, cai2016unified, lin2017focal}. 
The most notable embodiment of this framework is Faster-RCNN \citep{ren2015faster}, which was built upon the earlier Fast RCNN model \citep{girshick2015fast}. 
Deviating from the selective search methods utilized in Fast RCNN, Faster RCNN employs CNNs to generate region proposals via an efficient Region Proposal Network (RPN). The features from the final shared convolutional layer are then harnessed for both RPN's region proposal task and Fast RCNN's region classification task. In this study, we use Faster RCNN as one of the detectors in our semi-supervised framework.

\subsubsection{Anchor-free detectors}
While anchor-based detectors have demonstrated impressive outcomes, their application to novel datasets necessitates expertise in tuning hyperparameters \citep{jiao2019survey} associated with anchor boxes. This constraint limits the adaptability of these detectors to new datasets or environments \citep{zhang2020bridging}. Furthermore, anchor-based approaches are often proved to be computationally expensive for current mobile/edge devices used in agricultural applications, which typically have constrained storage and computational capacity.
Alternatively, these limitations are addressed in anchor-free detectors by getting rid of the need for pre-defined anchor boxes in detection models.
These methods can directly predict class probabilities and bounding box offsets from full images using a single feed-forward CNN without necessitating the generation of region proposals or subsequent classification/feature resampling, thereby encapsulating all computation within a single network \citep{liu2020deep}. YOLO (\underline{Y}ou \underline{O}nly \underline{L}ook \underline{O}nce) \citep{redmon2016you}, one of the most representative one-stage detectors, transforms the task of object detection into a regression problem by directly mapping image pixels to spatially separated bounding boxes and corresponding class probabilities. YOLO is designed for speed, capable of operating in real-time at 45 frames per second (FPS) by eliminating the region proposal generation process. On the other hand, FCOS (Fully Convolutional One-Stage Object Detection, \cite{tian2019fcos}) is an anchor box-free and proposal-free one-stage object detector. By eliminating the anchor box designs, FCOS avoids the complicated computation related to anchor boxes such as calculating overlapping during training and all hyper-parameters related to anchor boxes. In this study, FCOS serves as one of our base object detection models, chosen for its accessibility and extensive adoption within the field as evidenced by previous research \citep{li2021generalized, zhang2020bridging}.

\subsection{Semi-supervised Learning}
Semi-supervised learning, a form of label-efficient learning, leverages unlabeled samples to augment the learning process \citep{van2020survey, li2023label}. 
Most existing semi-supervised learning works \citep{tarvainen2017mean, berthelot2019mixmatch, xie2020self, sohn2020fixmatch, xu2021end} can be categorized into consistency regularization where the prediction is consistent with different perturbations, and self-training that involves an iterative update process. 

\begin{figure*}[!ht]
  \centering
  \includegraphics[width=0.55\textwidth]{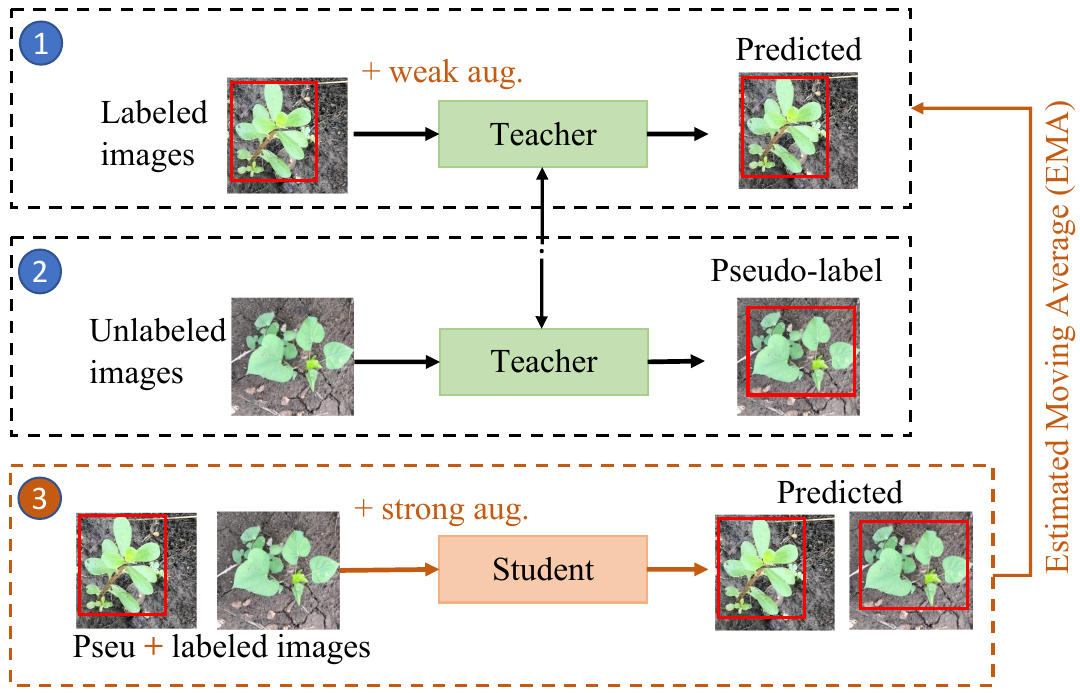}
  \caption{Pipeline of the proposed semi-supervised weed detection framework.}
  \label{fig: teacher_student}
\end{figure*}

The teacher-student framework is one of the mainstream ways for semi-supervised object detection \citep{sohn2020fixmatch, xu2021end, liu2021unbiased, li2022rethinking, chen2022label} using the self-training approach, which is illustrated in Figure~\ref{fig: teacher_student}.
Initially, a ``teacher'' model is trained on the labeled samples using supervised learning. This trained ``teacher'' model is duplicated into a "student" model and then employed to generate pseudo-labels for the unlabeled samples. Subsequently, a mixture of the most confidently selected pseudo-labeled samples and the original labeled samples are utilized to train a ``student'' model. Subsequently, the ``teacher'' model is updated with the ``student'' model using an Estimated Moving Average (EMA) strategy \citep{tarvainen2017mean} according to the following equation:
\begin{equation} \label{eqn:ema}
\centering
\theta_{\text{teacher}} = \alpha \cdot \theta_{\text{teacher}} + (1-\alpha) \cdot \theta_{\text{student}},   
\end{equation}
where $\theta_{\text{teacher}}$ and $\theta_{\text{student}}$ represent the parameters of the ``teacher'' and ``student'' models, respectively. The factor $\alpha$ determines the extent of the update. An $\alpha$ of 1 retains the original ``teacher'' model parameters, while an $\alpha$ of 0 fully replaces the "teacher'' model with the ``student'' model. In this study, we use cross-validations and find that $\alpha=0.99$ is the optimal choice for the designed semi-supervised learning framework. The EMA strategy serves as a crucial mechanism to reduce variance \citep{tarvainen2017mean}. We apply weak augmentation approaches (e.g., horizontal flip, multi-scale training with a shorter size range [400, 1200] and scale jittering) to the Student learning process and strong augmentation methods (e.g., randomly added gray scale, Gasussian blur, cutout patches \citep{devries2017improved}) to the Teacher learning processes, respectively, to enhance the performance during training process \citep{xu2021end, xie2020self}.
Figure \ref{fig: teacher_student} provides a visual representation of the described process.

This iterative process (steps 1-3) is repeated until the model achieves satisfactory performance. Upon completion of the model training, the ``student'' model is discarded, and only the ``teacher'' model is retained for inference. The versatility of self-training methods allows them to be integrated with any supervised learning-based approach, including one-stage and two-stage object detectors. In this study, we employ a self-training-based semi-supervised learning framework and assess two representative object detectors, Faster RCNN \citep{ren2015faster} and FCOS \citep{tian2019fcos}.

\subsubsection{Pseudo-labeling on Detectors}
It is important to obtain the most confident and accurate pseudo-labels in semi-supervised learning. Existing works \citep{liu2021unbiased, sohn2020simple, zhou2021instant} exploit the pseudo-labeling method to address semi-supervised object detection. The majority of existing works concentrated on anchor-based detectors. Our focus, however, lies in introducing the generalization approach for both anchor-free and anchor-based detectors, drawing inspiration from \citep{liu2021unbiased, liu2022unbiased}. 


We take the widely used FCOS model \citep{tian2019fcos} as an example for demonstrating the semi-supervised object detection tasks. 
FCOS comprises three prediction branches, classifier, centerness, and regressor, where the centerness score/branch dominates the bounding boxes score. 
However, the reliability of centerness scores in distinguishing foreground instances is questionable, particularly under conditions of limited label availability, as there is no supervision mechanism to suppress the centerness score for background instances within the centerness branch \citep{li2020generalized, liu2022unbiased}.
Consequently, although the centerness branch improves the anchor-free detector performance for the supervised training, it proves ineffective or even counterproductive for semi-supervised training scenarios \citep{li2020generalized, liu2022unbiased}. To address this issue, our approach prioritizes pseudo-boxes based solely on classification scores \citep{liu2022unbiased}.
The classifier is trained with the hard labels (i.e., one-hot vector) with the box localization weighting. Finally, we use the standard label assignment method instead of center-sampling, which designates all elements within the bounding boxes as foreground and everything outside as background.

\subsubsection{Unsupervised Regression Loss}
Confidence thresholding has proven effective in prior studies \citep{tarvainen2017mean, liu2021unbiased, sohn2020simple}. 
However, depending solely on box confidence is insufficient for effectively eliminating misleading instances in box regression,  since the ``teacher'' may still provide a contradictory regression to the ground-truth direction \citep{chen2017learning, saputra2019distilling}. 
To address this challenge, we categorize the pseudo-labels into two groups: beneficial instances and misleading instances. We then leverage the relative prediction information between the Student and the ``teacher'' to identify beneficial instances and filter out misleading ones during the training of the regression branch.
We define the unsupervised regression loss by selecting beneficial instances where the Teacher exhibits lower localization uncertainty than the Student by a margin of $\sigma$:
\begin{equation} \label{eqn:ap}
L_{reg}^{unsup} = \left\{\begin{matrix}
\sum _{i}\left \| \tilde{d_{t}^{i}} - \tilde{d_{s}^{i}} \right \|, \mathrm{if} \ \delta _{t}^{i} + \sigma \leqslant \delta _{s}^{i} \\ 
\ 0, \ \ \ \ \ \ \ \ \ \ \ \mathrm{otherwise}.
\end{matrix}\right.
\end{equation}
The parameter $\sigma \geqslant 0$ represents a margin between the localization uncertainties of the Teacher and the ''student'', where the localization uncertainty is loosely associated with the deviation from the ground-truth labels.
Specifically, $\delta _{t}^{i}$ represents the Teacher's localization uncertainty, while $\delta _{s}^{i}$ represents the Student’s localization uncertainty. Furthermore, $\tilde{d_{t}^{i}}$ and $\tilde{d_{s}^{i}}$ 
are the regression predictions for ''teacher'' and ''student'', respectively. For more details of the design for the unsupervised regression loss, please refer to \cite{liu2022unbiased}.

\subsection{Performance Evaluation Metrics}
In this evaluation, we rely on Average Precision (AP) as a primary metric, a measure derived from precision (P) and recall (R). AP summarizes the P(R) Curve to one scalar value.
However, since AP is traditionally evaluated for each object category separately, we employ the mean Average Precision (mAP) metric \citep{liu2020deep} to provide a comprehensive assessment across all object categories. The mAP is calculated as the average of AP scores over all object categories, and both AP and mAP are determined using the following equations:
\begin{equation} \label{eqn:ap}
AP = \int_0^1 P(R)dR,   
\end{equation}

\begin{equation} \label{eqn:map}
mAP = \frac{1}{n} \sum_{i=1}^n {AP}_i,   
\end{equation}
where $n$ represents the number of weed classes, and mAP signifies the average AP across these classes. A higher area under the Precision-Recall (PR) curve indicates improved object detection accuracy. 
Moreover, we consider mAP@[0.5:0.95], reflecting the mean average precision across IoU thresholds ranging from 0.5 to 0.95. These metrics collectively offer a representative evaluation of the model's performance across varying detection thresholds, ensuring a comprehensive understanding of its object detection capabilities.

\subsection{Experimental Setups}
In the process of model development and evaluation, the cotton weed dataset is partitioned into three subsets randomly. Specifically,
for a comprehensive evaluation, the CottonWeedDet3 dataset is randomly partitioned into training, validation, and testing sets following a ratio of 65\%:20\%:15\%, resulting in subsets comprising 550, 170, and 128 images. 
Similarly, the CottonWeedDet12 dataset is also divided into training, validation, and testing subsets, with a distribution ratio of $65\%$, $20\%$, and $15\%$, respectively. This results in subsets comprising 3670, 1130, and 848 images. The validation set is used to select the optimal trained model, while the test set is utilized to evaluate the model's performance. 


To expedite the model training process, we leverage transfer learning \citep{zhuang2020comprehensive} for all object detectors backbone, fine-tuning them with pre-trained weights obtained from the ImageNet dataset \citep{deng2009imagenet}. 
%
The model was implemented based on Detectron2 \citep{wu2019detectron2}. All models underwent training for 80k iterations, a duration deemed sufficient for effective modeling of the weed data. Stochastic Gradient Descent (SGD) is adopted as the optimizer, maintaining a momentum of 0.9 throughout the training process. The learning rate is selected as 0.01, and each batch contains 4 labeled images and 4 unlabeled images. We adopt the weak augmentation (horizontal flip, multi-scale training with a shorter size range [400, 1200] and scale jittering) for the Student,  and randomly add gray scale, Gasussian blur, cutout patches \citep{devries2017improved}, and color jittering as the strong augmentation for the Teacher. 
The computational setup includes a server running Ubuntu 20.04, equipped with two Geforce RTX 2080Ti GPUs, each with 12 GB of memory, ensuring efficient model training and testing.

\begin{figure*}[!ht]
    \centering
    \subfloat{{\includegraphics[width=.472\linewidth]{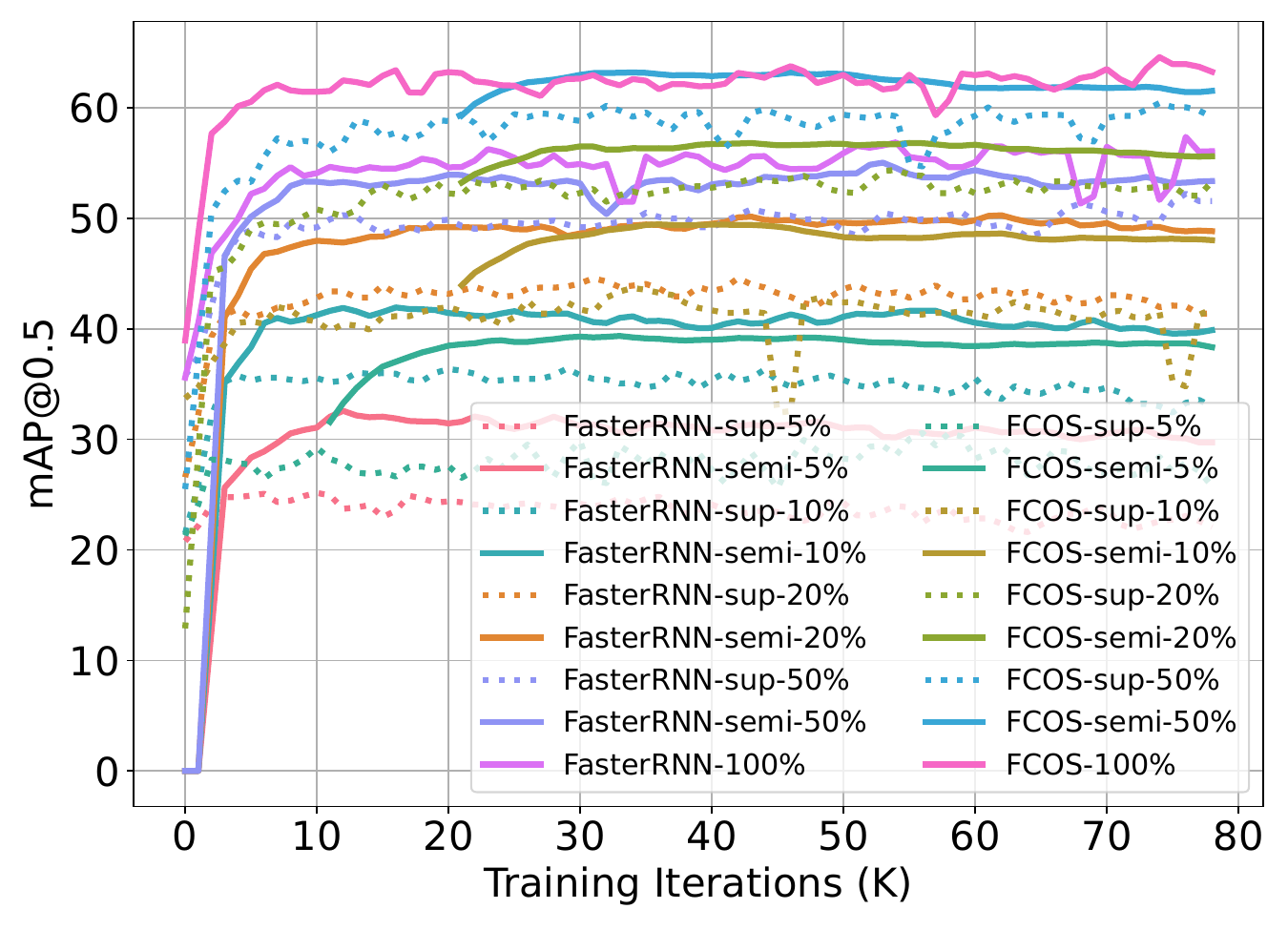} }}%
    \qquad
    \subfloat{{\includegraphics[width=.472\linewidth]{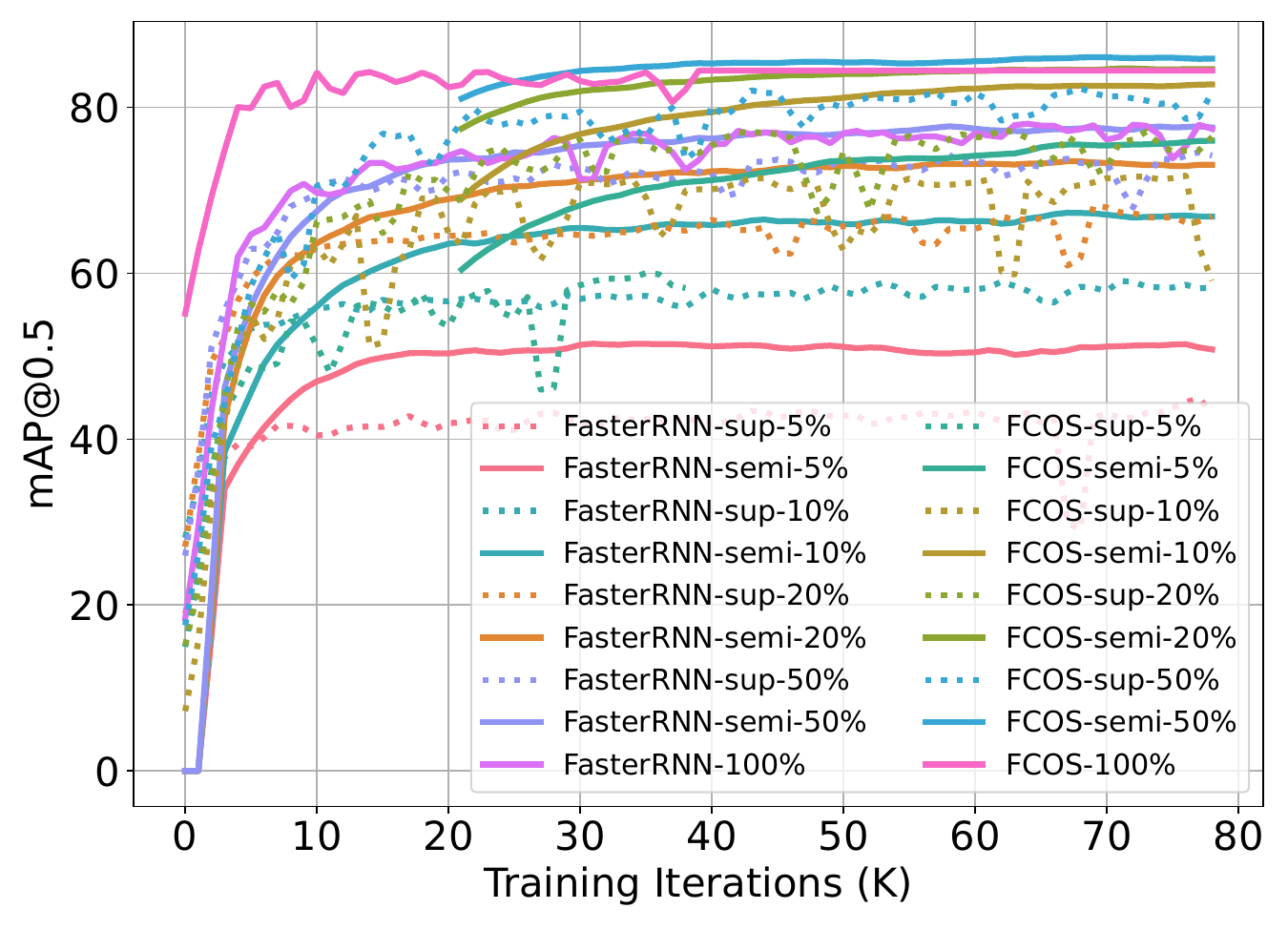}}}
    \caption{Training curves for FCOS and Faster RCNN with different proportions of labeled samples for two cotton weed datasets: CottonWeedDet3 (left) and CottonWeedDet12 (right).}
    \label{fig:train_acc}
\end{figure*}

\section{Experimental Results}
\label{sec: results}
In this section, we first evaluate the performance of various object detectors within the context of a semi-supervised learning framework. Subsequently, we will delve into a detailed analysis of the performance exhibited by individual weed classes.

\subsection{Semi-supervised Object Detector Comparison}
Figure~\ref{fig:train_acc} illustrates the training curves for FCOS and Faster RCNN, utilizing various proportions of labeled samples on the two cotton weed datasets: CottonWeedDet3 and CottonWeedDet12. We evaluate each algorithm in both supervised and semi-supervised learning contexts. For example, the configuration represented as Faster RCNN-sup-5\% refers to the Faster RCNN trained with supervised learning using 5\% of labeled samples. Conversely, Faster RCNN-semi-5\% is the same detector trained with semi-supervised learning using 5\% of the labeled samples and 95\% of the unlabeled samples.

It is evident from the results that semi-supervised learning outperforms its supervised counterparts on both datasets, given the exploitation of a large volume of unlabeled samples to bolster the training process. As an example, Faster RCNN-semi-5\% achieves superior training performance compared to Faster RCNN-sup-5\%. Moreover, it is noteworthy that FCOS-semi-50\% manages to attain performance comparable to that of FCOS-100\% (where all samples are labeled) on the CottonWeedDet3 dataset. FCOS-semi-50\% even surpasses FCOS-100\% on the CottonWeedDet12 dataset, suggesting that with only half the labeling effort, we can achieve improved performance, which also showcases that semi-supervised learning can be more robust compared with the supervised learning \citep{liu2021self}. 

Tables~\ref{tab:cottonweeddet3} and \ref{tab:cottonweeddet12} summarize the test performance (measured by mAP@[0.5:0.95]) comparison between the supervised and semi-supervised learning approaches based on the Faster-RCNN and FCOS models on the CottonWeedDet3 and CottonWeedDet12 datasets, respectively. Across both datasets, FCOS consistently outperforms Faster-RCNN in both the semi-supervised and supervised learning contexts. These findings are in agreement with the observations drawn from the training curves illustrated in Fig.~\ref{fig:train_acc}. For any given proportion of labeled samples, the semi-supervised learning approaches are found to enhance the test performance. For instance, on the CottonWeedDet3 dataset, the Faster RCNN model using a semi-supervised learning approach attains 86.70\% and 93.73\% of the performance of its supervised approach with only 20\% and 50\% of the samples labeled, respectively. Furthermore, it is worth highlighting that on the CottonWeedDet12 dataset, the FCOS model trained using semi-supervised learning with only 50\% of labeled samples outperforms the test performance of the fully supervised approach, which uses 100\% of the samples manually labeled. That is because semi-supervised learning can effectively leverage the vast amount of unlabeled samples, which may capture the inherent distribution of the data better than a limited set of labeled samples. 

\begin{figure}[!ht]
  \centering
  \includegraphics[width=0.48\textwidth]{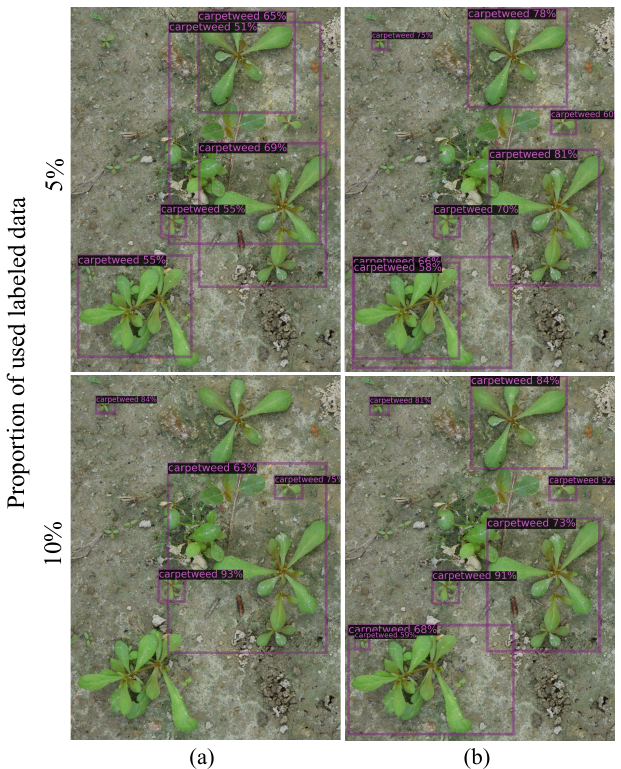}
  \caption{Examples of images annotated with ground truth labels (a) and predicted labels (b) using semi-supervised FOCS for CottonWeedDet3.}
  \label{fig:weed3_vis}
\end{figure}

\begin{figure*}[!ht]
  \centering
  \includegraphics[width=0.8\textwidth]{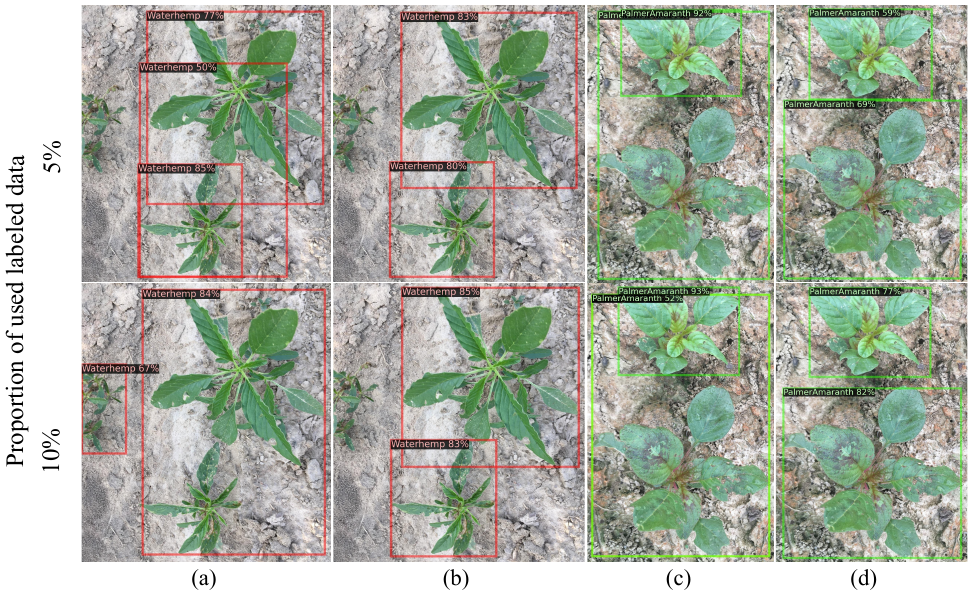}
  \caption{Comparing method results on CottonWeedDet12: (a) and (c) - supervised baseline, (b) and (d) - semi-supervised FCOS.}
  \label{fig:weed12_vis}
\end{figure*}

Figures~\ref{fig:weed3_vis} and \ref{fig:weed12_vis} show selected images predicted using both supervised and semi-supervised FCOS for CottonWeedDet3 and CottonWeedDet12, respectively. In both figures, only 5\% and 10\% of labeled samples are utilized for training. Remarkably, the semi-supervised FOCS exhibits visually compelling predictions, especially for images featuring diverse and/or cluttered backgrounds, as well as those with densely populated weed instances. Notably, the semi-supervised learning approach demonstrates superior performance compared to the supervised learning approach. For instance, in Figure~\ref{fig:weed3_vis}, the semi-supervised FOCS with 5\% labeled samples produces better predictions than the supervised learning approach with only 5\% labeled samples. This underscores the ability of semi-supervised learning to leverage valuable information from a large volume of unlabeled data.

\begin{table*}[!ht]
\renewcommand{\arraystretch}{1.4}
\centering
\caption{Testing performance (mAP@[0.5:0.95]) comparison between the supervised and semi-supervised based on Faster-RCNN and FCOS models on the CottonWeedDet3 dataset.}
\label{tab:cottonweeddet3}
\resizebox{0.58 \textwidth}{!}{%
\begin{tabular}{ccccccc}
\hline \hline
\multirow{2}{*}{Algorithms}  & \multirow{2}{*}{Supervision type} & \multicolumn{5}{c}{Proportion of labeled data for training}                                                                                            \\ \cline{3-7} 
                             &                                   & \multicolumn{1}{c}{5\%}   & \multicolumn{1}{c}{10\%}  & \multicolumn{1}{c}{20\%}  & \multicolumn{1}{c}{50\%}  & 100\%                  \\ \hline 
\multirow{2}{*}{Faster-RCNN} & Supervised                        & \multicolumn{1}{c}{21.14} & \multicolumn{1}{c}{31.90} & \multicolumn{1}{c}{42.65} & \multicolumn{1}{c}{50.51} & \multirow{2}{*}{56.75} \\ \cline{2-6}
                             & Semi-supervised                   & \multicolumn{1}{c}{29.33} & \multicolumn{1}{c}{40.17} & \multicolumn{1}{c}{49.20} & \multicolumn{1}{c}{53.19} &                        \\ \hline
\multirow{2}{*}{FCOS}        & Supervised                        & \multicolumn{1}{c}{27.37} & \multicolumn{1}{c}{42.20} & \multicolumn{1}{c}{52.42} & \multicolumn{1}{c}{59.84} & \multirow{2}{*}{62.80} \\ \cline{2-6}
                             & Semi-supervised                   & \multicolumn{1}{c}{38.17} & \multicolumn{1}{c}{47.93} & \multicolumn{1}{c}{55.79} & \multicolumn{1}{c}{61.32} &                        \\ \hline \hline
\end{tabular}
}
\end{table*}

\begin{table*}[!ht]
\renewcommand{\arraystretch}{1.4}
\centering
\caption{Testing performance (mAP@[0.5:0.95]) comparison between the supervised and semi-supervised based on Faster RCNN and FCOS models and CottonWeedDet12 dataset.}
\label{tab:cottonweeddet12}
\resizebox{0.58 \textwidth}{!}{%
\begin{tabular}{llccccc}
\hline 
\multirow{2}{*}{Algorithms}  & \multirow{2}{*}{Supervision type} & \multicolumn{5}{c}{Proportion of labeled data for training}                                                                                            \\ \cline{3-7} 
                             &                                   & \multicolumn{1}{c}{5\%}   & \multicolumn{1}{c}{10\%}  & \multicolumn{1}{c}{20\%}  & \multicolumn{1}{c}{50\%}  & 100\%                  \\ \hline 
\multirow{2}{*}{Faster-RCNN} & Supervised                        & \multicolumn{1}{c}{45.02} & \multicolumn{1}{c}{61.18} & \multicolumn{1}{c}{68.29} & \multicolumn{1}{c}{75.97} & \multirow{2}{*}{80.47} \\ \cline{2-6}
                             & Semi-supervised                   & \multicolumn{1}{c}{53.08} & \multicolumn{1}{c}{70.21} & \multicolumn{1}{c}{75.15} & \multicolumn{1}{c}{78.83} &                        \\ \hline
\multirow{2}{*}{FCOS}        & Supervised                        & \multicolumn{1}{c}{62.28} & \multicolumn{1}{c}{72.99} & \multicolumn{1}{c}{79.14} & \multicolumn{1}{c}{83.87} & \multirow{2}{*}{86.47} \\ \cline{2-6}
                             & Semi-supervised                   & \multicolumn{1}{c}{76.91} & \multicolumn{1}{c}{83.43} & \multicolumn{1}{c}{85.28} & \multicolumn{1}{c}{87.26} &                        \\ \hline  
\end{tabular}
}
\end{table*}

\subsection{Class-specific Performance}
Tables~\ref{tab:weed_performance3} and \ref{tab:weed_performance12} present the class-specific performance of the FCOS model on the CottonWeedDet3 and CottonWeedDet12 datasets, respectively. The instance count reflects the number of bounding boxes associated with each weed category within the test images. It is evident that the CottonWeedDet12 dataset exhibits a considerable imbalance, as indicated by the significantly uneven distribution of instances across various weed classes. 

On the CottonWeedDet3 dataset, the semi-supervised learning approaches demonstrate promising performance. Notably, the semi-supervised model trained with 50\% of the labeled samples surpasses the performance of the fully supervised learning model, particularly for Palmer Amaranth weeds. However, the detection accuracy for Carpetweed remains relatively low, attributed to its small size which poses an inherent challenge for recognition. A similar trend is observed in the performance metrics presented in Tables~\ref{tab:weed_performance12} for the CottonWeedDet12 dataset.

Remarkably, on the CottonWeedDet12 dataset, the semi-supervised FCOS model trained with 50\% and 20\% of labeled samples outperforms the fully supervised model for 8 out of 12 and 6 out of 12 weed classes, respectively. Impressively, for the top 3 minority weed classes — Cutleaf Groundcherry, Goosegrass, and Sicklepod — the FCOS model delivers superior performance even with only 50\% of the labeling costs compared to the supervised learning approach. This underscores the potential of semi-supervised learning models to effectively address class imbalance and provide superior performance even with fewer labeled samples.

\begin{table*}[!ht]
\renewcommand{\arraystretch}{1.4}
\centering
\caption{Test performance (mAP@[0.5:0.95]) on a specific category of weeds on CottonWeedDet3}
\label{tab:weed_performance3}
\resizebox{0.58 \textwidth}{!}{%
\begin{tabular}{lcccccc}
\hline
\multirow{2}{*}{Weeds} & \multirow{2}{*}{\# of instances} & \multicolumn{5}{c}{Proportion of labeled data for training}                                                                           \\ \cline{3-7} 
                       &                                  & \multicolumn{1}{c}{5\%}   & \multicolumn{1}{c}{10\%}  & \multicolumn{1}{c}{20\%}  & \multicolumn{1}{c}{50\%}  & 100\% \\ \hline \hline
PalmerAmaranth         & 100                              & \multicolumn{1}{c}{48.33} & \multicolumn{1}{c}{52.18} & \multicolumn{1}{c}{60.82} & \multicolumn{1}{c}{63.25} & 62.70 \\ 
MorningGlory           & 101                              & \multicolumn{1}{c}{46.82} & \multicolumn{1}{c}{55.47} & \multicolumn{1}{c}{63.56} & \multicolumn{1}{c}{65.97} & 70.83 \\
Carpetweed             & 93                               & \multicolumn{1}{c}{19.38} & \multicolumn{1}{c}{36.14} & \multicolumn{1}{c}{42.99} & \multicolumn{1}{c}{54.73} & 54.87 \\ \hline \hline
\end{tabular}
}
\end{table*}

\begin{table*}[!ht]
\renewcommand{\arraystretch}{1.4}
\centering
\caption{Test performance on the specific category of weeds on CottonWeedDet12}
\label{tab:weed_performance12}
\resizebox{0.58 \textwidth}{!}{%
\begin{tabular}{lcccccc}
\hline  \hline
\multirow{2}{*}{Weeds} & \multirow{2}{*}{\# of instances} & \multicolumn{5}{c}{Proportion of labeled data for training}                                                                                \\ \cline{3-7} 
                       &                           & \multicolumn{1}{c}{5\%}    & \multicolumn{1}{c}{10\%}   & \multicolumn{1}{c}{20\%}   & \multicolumn{1}{c}{50\%}   & 100\%  \\ \hline  
Waterhemp              & 352                       & \multicolumn{1}{c}{85.25} & \multicolumn{1}{c}{86}     & \multicolumn{1}{c}{88.52}  & \multicolumn{1}{c}{89.78} & 88.10 \\ 
MorningGlory           & 201                       & \multicolumn{1}{c}{83.82} & \multicolumn{1}{c}{85.75} & \multicolumn{1}{c}{87.40} & \multicolumn{1}{c}{89.71} & 88.03 \\ 
Purslane               & 161                       & \multicolumn{1}{c}{74.89} & \multicolumn{1}{c}{78.58} & \multicolumn{1}{c}{80.10} & \multicolumn{1}{c}{81.40} & 83.68 \\ 
SpottedSpurge          & 122                       & \multicolumn{1}{c}{77.12} & \multicolumn{1}{c}{81.49} & \multicolumn{1}{c}{83.59} & \multicolumn{1}{c}{85.79} & 82.78 \\ 
Carpetweed             & 137                       & \multicolumn{1}{c}{63.04} & \multicolumn{1}{c}{69.62} & \multicolumn{1}{c}{68.64} & \multicolumn{1}{c}{71.34} & 68.18  \\ 
Ragweed                & 144                       & \multicolumn{1}{c}{78.08} & \multicolumn{1}{c}{78.26} & \multicolumn{1}{c}{81.88} & \multicolumn{1}{c}{83.11} & 81.83 \\ 
Eclipta                & 117                       & \multicolumn{1}{c}{90.28} & \multicolumn{1}{c}{90.69} & \multicolumn{1}{c}{91.34} & \multicolumn{1}{c}{93.58} & 95.19 \\ 
PricklySida            & 60                        & \multicolumn{1}{c}{78.26} & \multicolumn{1}{c}{82.91} & \multicolumn{1}{c}{83.75} & \multicolumn{1}{c}{83.48} & 84.01 \\ 
PalmerAmaranth         & 42                        & \multicolumn{1}{c}{86.76} & \multicolumn{1}{c}{89.09} & \multicolumn{1}{c}{87.82} & \multicolumn{1}{c}{91.31} & 93.55  \\ 
Sicklepod              & 31                        & \multicolumn{1}{c}{94.27} & \multicolumn{1}{c}{96.56} & \multicolumn{1}{c}{97.09} & \multicolumn{1}{c}{97.01} & 96.43 \\ 
Goosegrass             & 31                        & \multicolumn{1}{c}{78.83} & \multicolumn{1}{c}{81.69}  & \multicolumn{1}{c}{85.58} & \multicolumn{1}{c}{90.02} & 85.31 \\ 
CutleafGroundcherry    & 15                        & \multicolumn{1}{c}{32.33} & \multicolumn{1}{c}{80.59} & \multicolumn{1}{c}{87.70} & \multicolumn{1}{c}{90.59} & 90.50 \\ \hline \hline
\end{tabular}
}
\end{table*}

\subsection{Comparative Analysis: Semi-Supervised Learning vs. Ground Truth Inaccuracies}
\begin{figure}[!ht]
  \centering
  \includegraphics[width=0.5\textwidth]{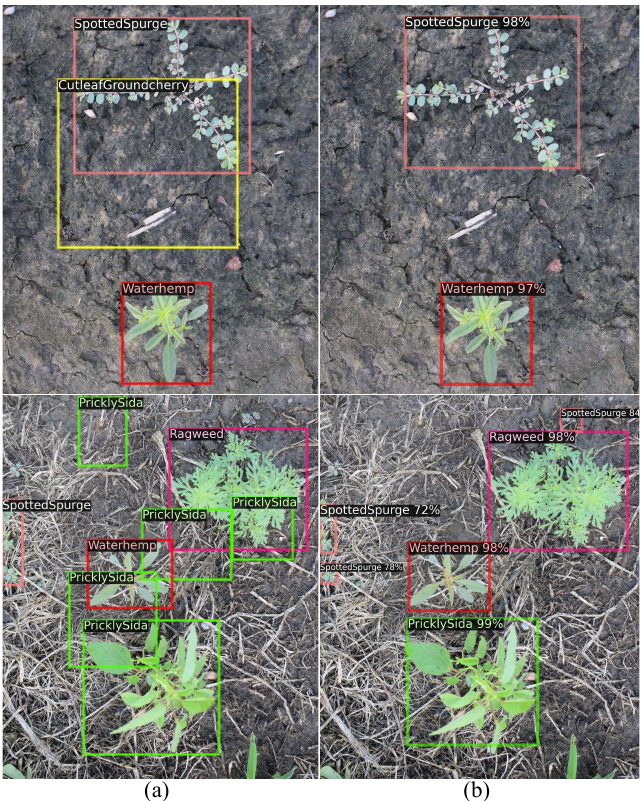}
  \caption{Image samples from CottonWeedDet12 with ground truth annotations (left) and predicted results with semi-supervised FCOS-10\% (right).}
  \label{fig:supsemi}
\end{figure}

In the preceding discussions, we demonstrate the remarkable performance improvement achieved by semi-supervised learning, even with a limited number of labeled samples, surpassing the results of traditional supervised learning approaches. In Figure~\ref{fig:supsemi}, we present image samples from CottonWeedDet12, showcasing both ground truth annotations and the predicted results obtained through the semi-supervised FCOS-10\%. Notably, a discernible observation is the presence of inaccuracies and mislabels in the ground truth annotations, highlighting the challenges associated with manual labeling by human experts, including instances of noise and incorrect labels. The application of a semi-supervised learning approach demonstrates to be a potent solution in mitigating the above challenges, and effectively enhancing accuracy and rectifying ground truth inaccuracies.

\section{Discussions}
\label{sec: disscussion}
The field of multi-class weed detection and localization remains largely unexplored in the existing literature \citep{dang2023yoloweeds, rai2023applications}. In the transition to the next-generation machine vision-based weeding systems, the focus is progressively shifting towards attaining higher precision and instituting weed-specific controls. Concurrently, the capability to differentiate between various weed species and identify individual weed instances emerges as an increasingly critical requirement within these vision tasks. While significant progress has been made in the development of DL-based weed detection \citep{dos2017weed, wang2019review, wu2021review, dang2022deepcottonweeds, dang2023yoloweeds}, these approaches typically rely heavily on expansive and manually-labeled image datasets, which makes these processes costly, prone to human error, and laboriously time-consuming. In our previous review on label-efficient learning in agriculture \citep{li2023label}, we presented various techniques aiming at reducing labeling costs and their respective applications in agricultural applications, including crop and weed management. Nevertheless, label-efficient technologies remain largely unexplored in the field of multi-class weed detection and localization. In this regard, this study stands as a unique contribution to the research community, specifically in the area of weed detection and control. By implementing semi-supervised learning, we introduce an innovative approach to alleviate the burden of labor-intensive labeling costs. Our evaluation includes both one-stage and two-stage object detectors on two open-source weed datasets, demonstrating that semi-supervised learning can significantly reduce labeling costs without substantial compromise  performance. Additionally, it can even generate enhanced performance metrics.

While this research provides valuable insights, it does acknowledge certain limitations that pave the way for potential future enhancements. Although the primary objective of this research is not to evaluate all deep learning-based object detectors for weed detection within the semi-supervised learning framework, there are indeed several high-performing object detectors that are not evaluated in this study. These include one-stage detectors such as SSD \citep{liu2016ssd}, RetinaNet \citep{lin2017focal}, EfficientDet \citep{tan2020efficientdet} and YOLO series \citep{terven2023comprehensive, dang2023yoloweeds}, as well as two-stage detectors like DINO \citep{zhang2022dino}, CenterNetv2 \citep{zhou2021probabilistic}, and RTMDet \citep{lyu2022rtmdet}. We intend to test and incorporate these models into our continually updated benchmark as we refine and improve the semi-supervised learning framework through future efforts.

In the scope of this study, we work under the assumption that all unlabeled samples are drawn from the same distribution as the labeled samples. It's important to acknowledge that unlabeled data might include instances from unknown or unseen classes, presenting a challenge commonly known as the open-set challenge \citep{chen2020semi}. This scenario may substantially compromise the efficacy of label-efficient learning. Consequently, we highlight a future investigation to delve into addressing out-of-distribution (OOD) issues, employing advanced sample-specific selection strategies. The aim is to identify and subsequently downplay the significance or utilization of OOD samples \citep{guo2020safe}. This planned exploration intends to enhance the generalization and robustness of our approach, ensuring its effectiveness in scenarios where the dataset contains samples from classes not encountered during the training phase, thereby contributing to a more resilient and versatile semi-supervised learning framework.

\section{Conclusion}
\label{sec: conclu}
In this study, we conducted an extensive evaluation of semi-supervised learning in the context of multi-class weed detection. Leveraging a set of labeled data alongside the unlabeled data for model training, our investigation focused on evaluating the efficacy of both one-stage and two-stage object detectors. The two datasets, CottonWeedDet3 and CottonWeedDet12, chosen for our study were meticulously curated to align with U.S. cotton production systems, ensuring the relevance of our findings to real-world agricultural scenarios. 
By leveraging semi-supervised learning, the labeling costs were significantly reduced,while only minimal impacts on the detection performance were observed. 
Additionally, by using the abundant unlabeled samples, the semi-supervised learning approach produced a more robust and accurate model, and it demonstrated the capability of mitigating noise and incorrect labels in the ground-truth annotations.
The outcomes underscore the potential of semi-supervised learning as a cost-effective and efficient alternative approach for developing agricultural applications, particularly those requiring extensive data annontations.

\section*{Authorship Contribution}
\textbf{Jiajia Li}: Conceptualization, Formal Analysis, Investigation, Methodology, Software, Validation, Visualization, Writing - original draft, Writing - review \& editing; \textbf{Dong Chen}: Conceptualization, Formal Analysis, Writing - original draft; \textbf{Xunyuan Yin}: Investigation, Writing - review \& editing; 
\textbf{Zhaojian Li}: Resources, Supervision, Writing - review \& editing.


\typeout{}
\bibliography{ref}
\end{document}